\documentclass[letterpaper]{article} 
\usepackage{aaai24}  
\usepackage{times}  
\usepackage{helvet}  
\usepackage{courier}  
\usepackage[hyphens]{url}  
\usepackage{graphicx} 
\usepackage{natbib}  
\usepackage{caption} 
\usepackage{algorithm}
\usepackage{algpseudocode}
\usepackage[dvipsnames]{xcolor}

\usepackage{amssymb}
\usepackage{mathtools}
\usepackage{amsmath}
\usepackage{amsthm}
\usepackage{cleveref}
\usepackage{pdfpages}

\usepackage{newfloat}
\usepackage{listings}
\DeclareCaptionStyle{ruled}{labelfont=normalfont,labelsep=colon,strut=off} 
\floatstyle{ruled}
\newfloat{listing}{tb}{lst}{}
\floatname{listing}{Listing}
\newcommand{\yantian}[1]{{\color{black} #1 \color{black}}}

\def\mund#1{\noindent{\bf #1}:}
\def\qund#1{\noindent{\bf #1}}

\title{Learning from Ambiguous Demonstrations with Self-Explanation Guided Reinforcement Learning}
\author {
    Yantian Zha\equalcontrib,
    Lin Guan\equalcontrib,
    Subbarao Kambhampati
}
\affiliations{
    Arizona State University\\
    \{yantian.zha,guanlin,rao\}@asu.edu
}

\usepackage{bibentry}

\begin{document}

\maketitle

\begin{abstract}
Our work aims at efficiently leveraging ambiguous demonstrations for the training of a reinforcement learning (RL) agent. An ambiguous demonstration can usually be interpreted in multiple ways, which severely hinders the RL agent from learning stably and efficiently. Since an optimal demonstration may also suffer from being ambiguous, previous works that combine RL and learning from demonstration (RLfD works) may not work well. Inspired by how humans handle such situations, we propose to use self-explanation (an agent generates explanations for itself) to recognize valuable high-level relational features as an interpretation of why a successful trajectory is successful. This way, the agent can leverage the explained important relations as guidance for its RL learning. Our main contribution is to propose the Self-Explanation for RL from Demonstrations (SERLfD) framework, which can overcome the limitations of existing RLfD works. Our experimental results show that an RLfD model can be improved by using our SERLfD framework in terms of training stability and performance. To foster further research in self-explanation-guided robot learning, we have made our demonstrations and code publicly accessible at \textcolor{blue}{https://github.com/YantianZha/SERLfD}. For a deeper understanding of our work, interested readers can refer to our arXiv version\footnote{\textcolor{blue}{https://arxiv.org/pdf/2110.05286.pdf\label{arxiv}}}, including an accompanying appendix.

\end{abstract}

\section{Introduction}
Reinforcement Learning from human visual demonstrations (RLfD) has gained prominence as an approach that improves the sampling efficiency of reinforcement learning (RL) by using the demonstrations to warm-start the learning process. RLfD offers advantages over pure imitation learning methods, such as overcoming distribution drift and limited adaptivity. As a driving force in the field of robotics, RLfD has shown great promise. However, it still suffers from sampling inefficiency when dealing with highly ambiguous demonstrations.

The challenges arise from the inherent ambiguity and noisy details present in human demonstrations, which can easily distract the learning process. This issue becomes even more vexing in tasks that involve both tacit and explicit knowledge components. For example, consider a task that requires understanding specific relations between objects (e.g., putting an object in a particular spatial relation with respect to another object) as well as a tacit motor demonstration of how the object is moved. In such cases, the ambiguity in the demonstration can be considerably reduced if the robot takes into account the possible space of explicit knowledge or symbolic relations that the human teacher could be interested in. To put it simply, imagine the difficulty of playing a game of ``dumb charades" when you are unsure of whether your friend is miming movie names or place names.
 
\begin{figure}[!t]
\centering
\includegraphics[width=8.5cm]{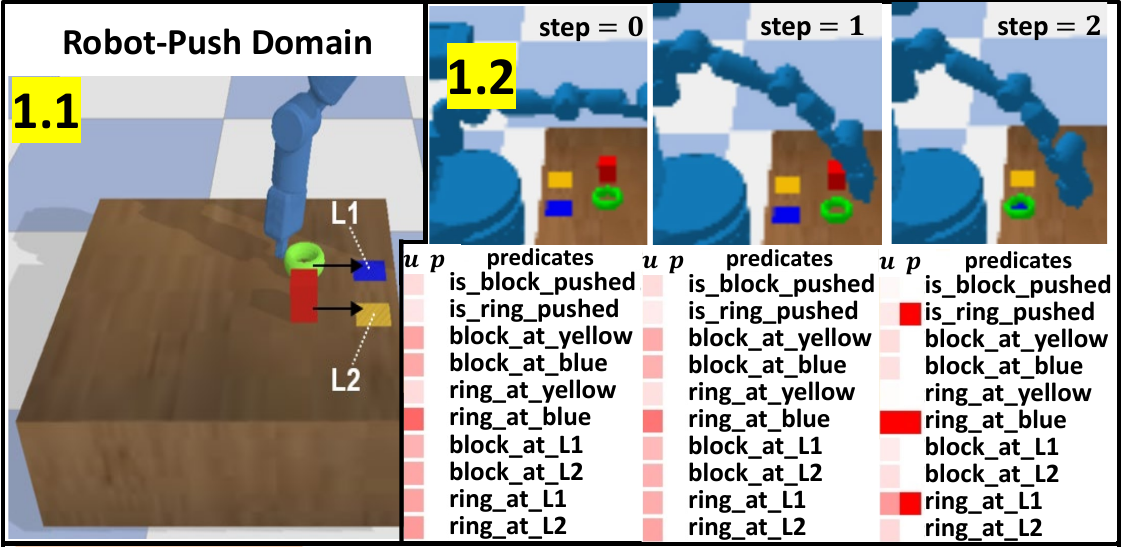} 
\caption{Fig \ref{fig:illustration}.1 shows the Robot-Push domain: There are two target regions which we can index as L1 and L2. L1 and L2 are also randomly assigned with the colors yellow-and-blue or blue-and-yellow in each episode. A human user demonstrates the task of pushing the ring and block into the blue and yellow region respectively. Fig \ref{fig:illustration}.2: a three-step robot execution with grounded predicates (\textbf{p}) and predicted self-explanations (\textbf{u}). }
\label{fig:illustration}
\end{figure} 

In this paper, we propose a framework inspired by cognitive psychology research, which suggests that humans enhance their learning by employing a self-explanation strategy. We introduce a self-explanation guided reinforcement learning approach where the robot attempts to ``self-explain" the potential relations over objects that the human teacher is considering. To support this, we establish a shared vocabulary of possible predicates in the domain, and the robot takes into account this space of predicate groundings when parsing a given demonstration. By doing so, we significantly reduce the ambiguity in the demonstration, effectively addressing the sample complexity issue.

To illustrate the challenges of learning from ambiguous demonstrations, we consider an everyday household robot scenario in Fig. \ref{fig:illustration}. In this scenario, a robot needs to push two components into different target regions, denoted as L1 and L2, which are distinguished by their colors, blue and yellow, respectively. Suppose a human teleoperates the robot to push a ring object into the blue L1 region and a block object into the yellow L2 region. This demonstration can be interpreted in two ways: either the robot should push the block into the yellow region and the ring into the blue region, or it should push the block into L2 and the ring into L1.

If humans were in the robot's position, they would employ a self-explanation strategy to learn from ambiguous demonstrations. Drawing on their background knowledge of important relations in the environment, they would form hypotheses about the relevant object relations and test these hypotheses by taking actions accordingly. If the initial hypothesis fails, they would adjust their explanations and try a different approach. In our work, the task of self-explaining involves identifying the task-relevant relations among all the domain relations at a given state, which aids in disambiguating demonstrations during the learning process.

We assume that the robot possesses human-related background knowledge, which includes relations and symbols, as well as classifiers for extracting relations in the domain. By incorporating such knowledge into the learning process, robots can benefit from the disambiguation capabilities of self-explanation. The key insight behind our approach is the explicit identification of task-relevant object-event relations, which can guide and improve the learning process.

However, learning self-explanations from non-expert humans in the absence of explicit labels presents a significant challenge. Human demonstrators provide non-symbolic and potentially ambiguous demonstrations without self-explanation labels. To tackle this challenge, we propose the Self-Explanation for Reinforcement Learning from Demonstration (SERLfD) framework. We adapt Generative Adversarial Inverse Reinforcement Learning (GAN-IRL) methods, which simultaneously train a generator (RL agent) and a discriminator to distinguish between demonstrations and generated trajectories. In our framework, we enhance the discriminator in GAN-IRL with a non-linear self-explanation predictor (SE-Net). The SE-Net predicts predicate utilities as self-explanations from raw data. Unlike GAN-IRL, our discriminator distinguishes between successful and failed trajectories, similar to the reflective nature of humans. We utilize the SE-Net to assist RL agents in maximizing environment rewards by augmenting states and/or rewards with self-explanations. As a result, our SERLfD algorithm combines the benefits of RLfD and GAN-IRL.

To the best of our knowledge, the proposed SERLfD framework pioneers the direction of learning self-explanations from agents' past experiences and human demonstrations to guide deep reinforcement learning. This opens up a broad space for future research. Our main contributions include: 1) Introducing the SERLfD framework that contrastively learns to identify task-relevant predicates while performing the task iteratively; 2) Extensive evaluation of the SERLfD framework with multiple RL agents and various ways of using self-explanations; and 3) Demonstrating the superiority of self-explanations over traditional RLfD and GAN-IRL methods in terms of learning stability and performance, even in challenging continuous control domains.

\section{Related Works} \label{sec:rw}
\mund{Imitation Learning from Ambiguous Demonstrations}
In the realm of learning from ambiguous demonstrations, only a handful of studies have explored the field of robot learning from demonstrations (LfD). Breazeal et al. \cite{breazeal2006using} and Bensch et al. \cite{bensch2010ambiguity} approach the problem by modeling demonstration ambiguity as differences in the intentions of humans and robots. Breazeal et al. \cite{breazeal2006using} propose a Bayesian inference framework that explicitly models human intention and belief, enabling the robot to identify conflicts, seek clarification from humans, and enhance its learning process. Bensch et al. \cite{bensch2010ambiguity} view ambiguity as a hypothesis space comprising multiple categories of concepts (e.g., colors, shapes) and address disambiguation by gradually reducing the hypothesis space through concept learning with additional demonstrations. Morales et al. \cite{morales2013ambiguity} tackle demonstration ambiguity by focusing on differences in demonstrated actions within the same or similar states. They employ clustering algorithms and propose a two-level clustering approach to categorize similar situations and ambiguous actions within each situation. Another approach by Brown et al. \cite{brown2020bayesian} leverages Bayesian optimization to infer reward uncertainty learned through an Inverse Reinforcement Learning (IRL) algorithm, demonstrating the potential for imitation learning from ambiguous demonstrations. However, the work primarily concentrates on the IRL aspect and evaluates its performance in simple domains. 
A distinct direction is taken by De et al. \cite{de2019causal}, who address the issue of distribution drift in imitation learning by mitigating causal misidentification resulting from ambiguous demonstrations. Although their work touches on RL, it does not fall under the RLfD paradigm since it does not utilize task rewards. In contrast, RLfD approaches, including our own, as highlighted by Rajeswaran et al. \cite{rajeswaran2017learning}, offer several advantages over LfD approaches: (1) they require fewer demonstrations and (2) they naturally handle distribution drift problems by directly incorporating task rewards into the learning process.

\mund{Deep RL from Demonstrations}
Numerous studies have explored the advantages of using RLfD frameworks. Salimans et al. \cite{salimans2018learning} utilize states in demonstrations as starting points for training Deep Reinforcement Learning (DRL) models with short-term interactions. This approach focuses exploration in the local region surrounding promising states from the demonstrations. Hester et al. \cite{hester2017deep} and Vecerik et al. \cite{vecerik2017leveraging} propose initializing the replay buffer with demonstrations to facilitate training a Deep Q-Network (DQN) \cite{hester2017deep} or a Deep Deterministic Policy Gradient (DDPG) network \cite{vecerik2017leveraging}. Pretraining neural network parameters with an imitation learning objective is advocated by Cruz et al. \cite{cruz2017pre}, Rajeswaran et al. \cite{rajeswaran2017learning}, and Pfeiffer et al. \cite{pfeiffer2018reinforced}, who directly use demonstrations for initialization. In addition to pretraining, auxiliary imitation learning losses can be constructed, as demonstrated by Vecerik et al. \cite{vecerik2017leveraging} and Nair et al. \cite{nair2018overcoming}. Gao et al. \cite{gao2018reinforcement} tackle the problem of RL from imperfect demonstrations, focusing on noisy and corrupted demonstrations. Their approach incorporates reward information to normalize the Q-function over actions. However, the issue of RL from ambiguous demonstrations addressed in our work is distinct, as these demonstrations may yield the highest reward and may not necessarily be corrupted but can confuse the learner. However, reinforcement learning from ambiguous demonstrations that we address is an orthogonal issue, as these demonstrations may yield the highest reward and may not necessarily be corrupted but can confuse the learner.

\vspace*{-3mm}
\section{Our Approach}
\subsection{Background}
\mund{Reinforcement Learning and Reward Shaping}
In our work, we consider a finite-horizon and discounted Markov decision process (MDP) model that can be learned by RL methods. We further assume that the reward function $r(s_t,a_t)$ is sparse, i.e., $r(s_t,a_t)=0$ in most of the states $s \in S$. Training RL agents in sparse rewards environments could be challenging due to the delayed training signals from effective feedback. One extensively adopted way to ameliorate such training is by adding reward shaping, which provides denser training signals so that the agent could obtain valuable feedback much sooner. However, we need to be careful at providing shaped rewards. As demonstrated in \cite{ng1999policy}, a poorly-designed reward shaping function may cause the converged optimal policy to shift as against the one under original rewards. \cite{ng1999policy} proves that potential-based reward shaping function, which follows the form in Eq. \ref{eq:rwd_shaping}, is the only class of reward shaping function that can guarantee the invariance of optimal policies.
\begin{equation} \label{eq:rwd_shaping}
    \hat{r}(s_t,a_t) = r(s_t,a_t) + \lambda\Phi(s_{t+1}) - \Phi(s_{t})
\end{equation} where $\hat{r}(s_t,a_t)$ and $r(s_t,a_t)$ denote the shaped and original reward respectively, $\lambda$ is an adjustment parameter, $\Phi$ denotes any real-valued function, and  $\lambda\Phi(s_{t+1}) - \Phi(s_{t})$ is the reward shaping term. Note that \textit{we set $\lambda$ to 1 in the rest of paper}.

\mund{Inverse Reinforcement Learning (IRL)}
IRL is about the problem that the reward function in an MDP model is unknown and needs to be discovered from expert demonstrations. Typically in an IRL framework, the unknown reward function is modeled by a parameterized function of certain features \cite{ziebart2008maximum}. The design of our SERLfD framework is based on GAN-IRL methods like the Guided Cost Learning (GCL \cite{finn2016guided}) optimized by Generative-Adversarial Networks (GAN-GCL \cite{finn2016connection}). GCL and GAN-GCL propose to couple IRL and RL together to do continuous control learning with unknown rewards. Such frameworks allow learning nonlinear rewards that help under complex and unknown dynamics (essentially a model-free IRL).



GAN-GCL integrates IRL and RL by viewing the RL model as a generator and the IRL model as a discriminator that are trained in GAN formulation. The IRL model provides rewards for the RL model. The RL model is trained to gradually shift its sampling distribution to match that of demonstrations. The IRL model is trained to distinguish sampled trajectories from demonstration trajectories, by using the binary cross-entropy loss in Eq. \ref{eq:gan_gcl}. Using such a discriminator to provide rewards for policy learning can solve their Imitation Learning problems.

\vspace*{-6mm}
\begin{gather}
    L_{irl}(D_\theta)=\mathbb{E}_{\tau \sim p}[-logD_\theta(\tau)]+\mathbb{E}_{\tau \sim q}[-log(1-D_\theta(\tau))] \label{eq:gan_gcl} 
\end{gather} where $D_\theta$ denotes a discriminator model. $\tau$ denotes a trajectory. The work Adversarial IRL \cite{fu2017learning} further proposes to replace trajectory $\tau$ with state-action pairs, which makes the training more stable:


\begin{equation} \label{eq:airl}
D_\theta(s, a) = \frac{exp\{\hat{r}_\theta(s,a)\}}{exp\{\hat{r}_\theta(s,a)\} + q(a|s)}
\end{equation} where $\hat{r}_\theta(s,a)$ denotes the estimation of reward feature on a pair of state and action. The $q(a|s)$ is a policy function that can be used to calculate the probability of taking the action $a$ at state $s$. We refer to this version of GAN-GCL as State-Action-GAN-GCL (\textbf{SA-GAN-GCL}).

Note that our SERLfD framework is essentially solving an RLfD problem as in \cite{vecerik2017leveraging,hester2017deep,salimans2018learning}. RLfD assumes that there exists some task reward signal from the environment. By populating the replay buffer with demonstrations, RLfD methods are able optimize the environment reward more efficiently than ordinary RL. In contrast, IRL problem assumes that there is no environment reward, so the agent has to simultaneously learn a reward function and a policy. As we will see later, although our self-explainer employs a similar computational framework to some IRL methods \cite{fu2017learning,finn2016connection}, we are not solving an IRL problem because the learned self-explanation (predicate utility weights) in SERLfD is used to \textit{augment original state and/or reward}, as a means to mitigate the negative effects of ambiguity in demonstrations.

\subsection{The SERLfD Framework} \label{sec:ap}

\begin{figure}[!h]
\centering
\includegraphics[width=1.\columnwidth]{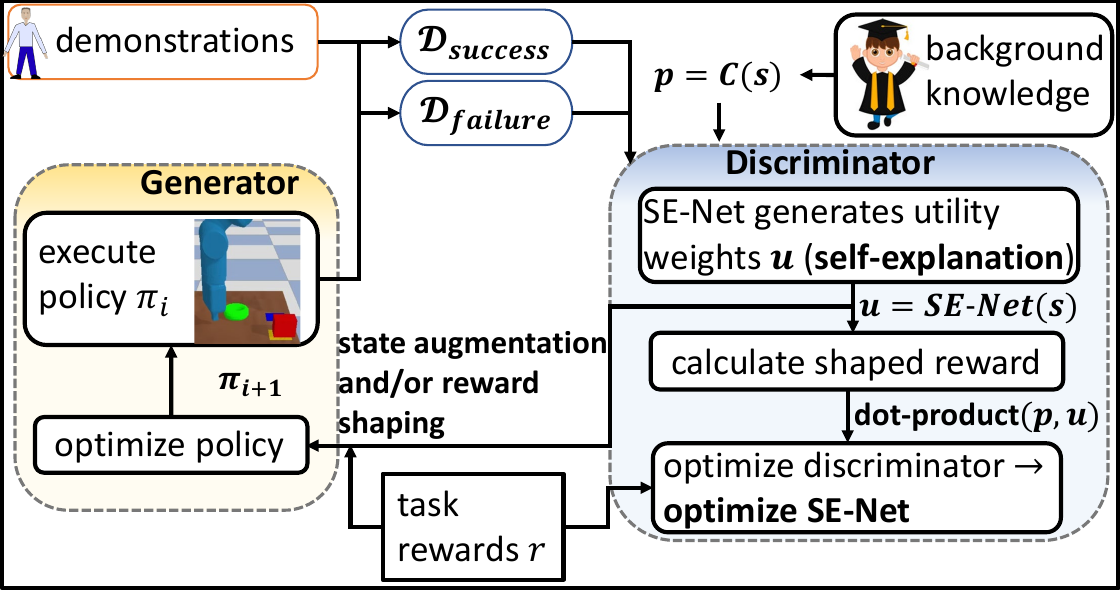} 
\caption{The SERLfD framework that couples learning Self-Explanations Networks (SE-Nets) and an RL agent. 
Roboticists provide predicates as human-related background domain knowledge to help robots disambiguate non-expert demonstrations for specific tasks. Buffers $\mathcal{D}_{success}$ stores successful experiences which include those from demonstrations. $\mathcal{D}_{failure}$ stores unsuccessful experiences. We train Self-Explanation Network (SE-Net) by inserting it into a Discriminator that distinguishes between successful and failed experiences in a Generative-Adversarial framework.}
\label{fig:SERLfD}
\vspace{-1em}

\end{figure}

The main idea behind our SERLfD framework is to support RLfD by simultaneously learning to identify which of the domain predicates are relevant to accomplishing a task at each step, i.e. to self-explain. SERLfD interleave the learning of two sub-tasks: 1) Train a generator (RL agent) by maximizing the accumulative environment rewards while being guided by self-explanations; and 2) Train the Self-Explainer by including it in a Discriminator to distinguish between successful and unsuccessful experiences. Fig. \ref{fig:SERLfD} illustrates
the main components of our approach.

\mund{Predicates} \label{sec:problem}
In our approach, we assume that robotics experts provide robots with task-agnostic domain knowledge in the form of human-understandable predicates. This knowledge is aimed at assisting robots in disambiguating non-expert human demonstrations. A predicate is a logical expression used to describe a relationship among objects. The objects involved in the predicate are referred to as its arguments. Take for instance a predicate ``pushed(obj)'' in the aforementioned robot-pushing domain, ``obj'' is an argument that can be grounded with ``block'' or ``ring''. When the argument is grounded, the predicate becomes a grounded predicate, e.g. ``pushed(block)''. In our work, we use a binary predicate variable to represent if a grounded predicate is satisfied or not. For example, ``pushed(block)'' can be represented by a binary predicate variable ``is\_block\_pushed''. When it is satisfied, we assign this binary predicate variable with value 1, otherwise -1. 

In our framework, we establish a collection of $N$ binary predicate variables, denoted as $P=\langle p_1,p_2,...,p_N \rangle$, which corresponds to a domain $D$ capable of supporting various tasks, where $p_i$ is the $i$-th binary predicate variable in the set $P$. We also assume to have a set of $N$ classifiers\footnote{We note that some robot LfD works rely on marker tracking to simplify the complexity of the vision part of their theories, e.g.  \cite{konidaris2012robot,niekum2012learning}. We take a cue from this and provide a procedural way to build each classifier. \yantian{For details, please consult Appendix.B in the arXiv version (see \cref{arxiv}).}} $C=\langle c_{p_1},c_{p_2}...,c_{p_N} \rangle$, where $c_{p_i}$ is the classifier that can determine whether the $i$-th predicate is satisfied given a state: $p_i = c_{p_i} (s) =$ -1 or 1.

\mund{Self-Explanation}
The self-explanation mechanism operates by training a SE-Net to predict a set of predicate utility weights, denoted as \textit{u}. These weights, estimated without relying on ground-truth self-explanations, serve to determine the relevance of predicates to the task at hand in a given state. Mathematically, we express this as: $u = \langle u_1, u_2, ..., u_N \rangle = SE\text{-}Net(s)$. Each predicate utility weight $u_i$ (where $1 \leq i \leq N$) represents a quantitative estimation of the utility of the corresponding predicate variable in elucidating the viability of a decision. The values encompassed within \textbf{\textit{u}} may accentuate certain predicate variables while diminishing the significance of others.

\mund{Self-Explanation Guided Learning of RL agent (Generator)} \label{sec:serl_rl_agent}
Since the predicates may provide meaningful abstraction of the environment states and may potentially speed up RL training, we append the predicate values $P$ to the environment state (RL state), resulting in the transformation $\langle s \rangle \rightarrow \langle s, p \rangle$. The SE-Net and RL agent mentioned earlier are trained iteratively. The yellow area in Fig. \ref{fig:SERLfD} illustrates the training process of the RL agent, leveraging self-explanations \textit{u} predicted by the SE-Net. Notably, the SE-Net learns a sophisticated non-linear function to identify task-relevant predicates within a continuous state, providing detailed guidance (\textit{u}) beyond mere numerical reward predictions. The self-explanations \textit{u} can be used to assist RL agents in two aspects, namely by forming shaping rewards (reward augmentation) or by concatenating them to RL states (state augmentation).

In a similar fashion to solving an RLfD problem, the objective of SERLfD is to leverage demonstrations for training the RL agent to maximize task (environment) rewards. To accomplish this, we utilize self-explanations to construct a reward shaping value that augments the rewards. The reward shaping value, denoted as $h_\theta(s,u)$, is computed as follows:
\begin{equation} \label{eq:dot-prod}
\begin{alignedat}{2}
h_\theta(s,u) = \sum \limits_{i=0}^{k-1} u_i \cdot  c_{p_i} (s)
\end{alignedat}
\end{equation}
where $h_\theta(s,u)$ is a scalar value used as the reward shaping value, $c_{p_i} (s)$ represents the application of the $i$-th predicate detector $c_{p_i}$ to obtain the grounded predicate value from the state $s$, and $k$ denotes the number of predicates. Then the shaped rewards $\hat{r}$ can be calculated as follows:
\begin{equation} \label{eq:serl_rwd}
\begin{split}
    &\hat{r}_\theta(s_t,u_t,a_t,s_{t+1},u_{t+1}) = r(s_t,a_t)+
    h_\theta(s_{t+1},u_{t+1})\\
    &-h_\theta(s_t,u_t) 
\end{split} \hspace{-5pt}
\end{equation} where the function $\hat{r}_\theta$ models the prediction of shaped rewards, denoted as $\hat{r}(s_t,a_t)$. The value of $h_\theta(s,u)$, computed in Eq. \ref{eq:dot-prod}, is included in this computation. The term $r(s_t,a_t)$ represents the task rewards in the RL environment. Eq. \ref{eq:serl_rwd} is formulated based on Eq. \ref{eq:rwd_shaping} and shares similarities with the prediction of shaped reward in work \cite{fu2017learning}. However, unlike \cite{fu2017learning}, our approach incorporates the task reward $r$ to train the agent to accomplish the task instead of solely imitating. 

The reward augmentation way of using self-explanations provides more direct learning signals for RL agents. However, we could also use self-explanations \textit{u} to augment RL states: $\langle s,p \rangle \rightarrow \langle s,p,u\rangle$. The benefit is that it provides more detailed guidance for RL agents. Depending on whether we augment task rewards with self-explanations, the learning objective of RL agents could be either maximize the accumulative rewards of the environment rewards $r$, or the augmented rewards $\hat{r}_\theta$ computed by using Eq. \ref{eq:serl_rwd}.

\mund{Training the Self-Explanation Network (SE-Net) in a Discriminator} \label{sec:senet}
To describe the training procedure for generating self-explanations $u$ and training the SE-Net, we follow these steps. First, we sample a batch of experiences and grounded predicate values from the buffers $\mathcal{D}_{success}$ and $\mathcal{D}_{failure}$, which store experiences from successful and unsuccessful trajectories. These samples serve as the input to the Discriminator, represented by the blue area in Fig. \ref{fig:SERLfD}. Since the SE-Net is integrated into the Discriminator, we can optimize a binary logistic regression loss \cite{king2008binary} to train the SE-Net. The specific neural network architecture of the SE-Net can be customized to match the characteristics of the state space. For instance, in experiments where the state space involves features such as object poses, a multilayer perceptron can be employed to implement the SE-Net.

Our Discriminator design aligns with the Generator in terms of utilizing self-explanations. The purpose of our self-explanation $u$ is to enhance the task rewards and facilitate faster learning of tasks by RL agents, rather than solely performing imitation learning from ambiguous demonstrations. Consequently, our Discriminator is trained to differentiate between successful and unsuccessful trajectories to identify task-relevant predicates as self-explanations. As depicted in Fig. \ref{fig:SERLfD}, we store a sampled trajectory into either $\mathcal{D}_{success}$ or $\mathcal{D}_{failure}$ depending on whether it accomplishes the task or not. The discriminator loss function $L_{SE}$, responsible for training the SE-Net, is defined in Eq. \ref{eq:l_se}.

\begin{algorithm*}
  \caption{The SERLfD Learning Algorithm}\label{alg:serlfd}
  \textbf{INPUT: A dataset of human demonstrations, an environment with reward function $r$, state space $S$, action space $A$, and all hyper-parameters} \vspace*{-5mm}\\
  \begin{algorithmic}[1] 
    \State Initialize the weights of an RL agent and a Self-Explainer (SE-Net)
    \State Initialize buffers $\mathcal{D}_{success}$ and $\mathcal{D}_{failure}$ for training the Self-Explainer and $D_{RL}$ for RL from Demonstrations as in the RLfD works \cite{hester2017deep,vecerik2017leveraging} 
    \State Store expert experiences into $\mathcal{D}_{success}$ and $\mathcal{D}_{RL}$. Pretrain the RL agent with experiences sampled from $\mathcal{D}_{RL}$ 
    \State Sample $K$ trajectories with a random policy and add them to $D_{RL}$. Also add successful and unsuccessful trajectories to $\mathcal{D}_{success}$ and $\mathcal{D}_{failure}$ respectively.
    
    \For{\texttt{$episode=1; episode \leq N; episode++$}} 
        \State Sample experiences (including grounded predicate values) from $\mathcal{D}_{success}$ and $\mathcal{D}_{failure}$ \Comment{\textbf{Train Self-Explainer}} 
        \State Compute utility weights $u$ and shaped reward prediction $\hat{r}_\theta(s_t,u_t,a_t,s_{t+1},u_{t+1})$ by using SE-Net, Eqs. \ref{eq:dot-prod} and \ref{eq:serl_rwd} 
        \State Update the SE-Net via binary cross entropy loss $L_{SE}$ (Eq. \ref{eq:l_se}) to distinguish successful experiences from unsuccessful experiences 
        \State Sample experiences with grounded predicate values from $\mathcal{D}_{RL}$ \Comment{\textbf{Train RL agent}}
        \State Run SE-Net on sampled states to obtain utility weights $u$ 
        \State Augment input states with grounded predicate values and utility weight values 
        \State Augment rewards with predicted shaped reward by using Eq. \ref{eq:serl_rwd} 
        \State Update RL agent with the augmented experiences 
        \State Use RL agent to sample a new trajectory and add it to $D_{RL}$. If the trajectory is successful, it would also be added to $\mathcal{D}_{success}$. Otherwise, it would also be added to $\mathcal{D}_{failure}$ \Comment{\textbf{Sample a new trajectory}}
    \EndFor \\
    \Return a trained RL agent and SE-Net 
  \end{algorithmic}
\end{algorithm*}


\vspace*{-2.5mm}

\begin{equation} \label{eq:l_se}
\begin{split}
    &L_{SE}(s_t,a_t) = \mathbb{E}_{(s,a) \sim \mathcal{D}_{success} }[-logD(\hat{r}_\theta(s_t,u_t,a_t,s_{t+1},\\
    &u_{t+1}))]+\mathbb{E}_{(s,a) \sim \mathcal{D}_{failure}}[-log(1-\\
    &D(\hat{r}_\theta(s_t,u_t,a_t,s_{t+1},u_{t+1})))] 
\end{split} \hspace{-18pt}
\end{equation} where the discriminator function $D(\cdot)$ is formulated in Eq. \ref{eq:airl}. The shaped reward $\hat{r}_\theta(s_t,u_t,a_t,s_{t+1},u_{t+1})$ is computed by using Eqs. \ref{eq:dot-prod} and \ref{eq:serl_rwd}. The predicate utility weights (self-explanation) $u$ in the function $\hat{r}(\cdot)$ is predicted by the Self-Explainer Network (SE-Net).





The SERLfD learning is summarized in \textbf{Algorithm. \ref{alg:serlfd}}.


\section{Evaluation} \label{sec:eval}



Since self-explanations should play a general role in improving an RL agent that is trained with ambiguous demonstrations, we evaluate our SERLfD framework in multiple domains and use different candidate deep RL models as the RL agent. We design our evaluation to answer the following questions: 1) Can SERLfD outperform RLfD? 2) What are promising ways of using self-explanations to guide an RL agents (i.e., state augmentation, reward shaping, or both)? 3) Do self-explanations play a general role in supporting an RL agent to learn from ambiguous demonstrations or self-explanations are only effective for certain of the RL models? 4) Since our SERLfD combines the benefits of RLfD and GAN-IRL, does our SERLfD outperform a state-of-the-art GAN-IRL for Imitation Learning method? 5) Does SERLfD help in both continuous and discrete domains?

\begin{figure*}[!t]
\centering
\includegraphics[width=\textwidth]{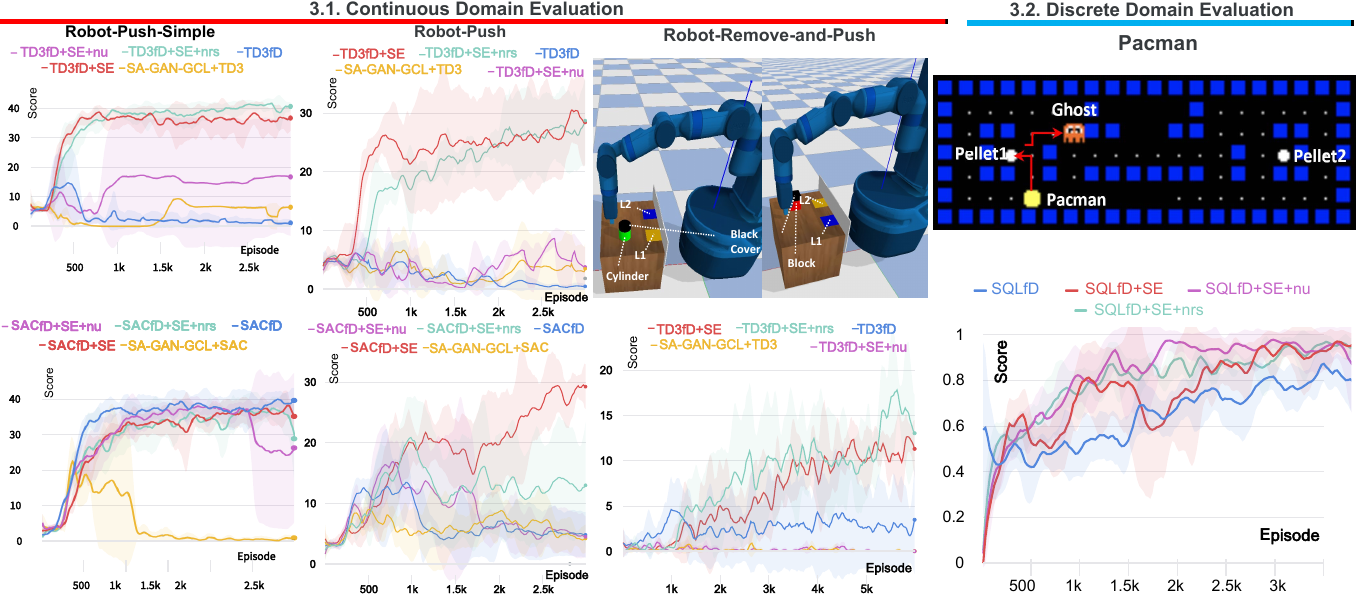}
\caption{Learning curves of training the \textcolor{blue}{baseline RLfD} agents (\textcolor{blue}{TD3fD/SACfD}), TD3fD/SACfD with SE-Nets that uses self-explanation to augment states (TD3fD/SACfD+SE+\textbf{nrs}), augment rewards (TD3fD/SACfD+SE+\textbf{nu}), or both (TD3fD/SACfD+SE), and an Imitation Learning agent built by using RL in the original SA-GAN-GCL framework \cite{fu2017learning}; The \textcolor{blue}{blue}, \textcolor{red}{red}, \textcolor{Aquamarine}{aqua}, \textcolor{Magenta}{magenta}, and \textcolor{Goldenrod}{gold} curves are the results of \textcolor{blue}{baseline TD3fD/SACfD}, \textcolor{red}{TD3fD/SACfD+SE}, \textcolor{Aquamarine}{TD3fD/SACfD+SE+\textbf{nrs}}, \textcolor{Magenta}{TD3fD/SACfD+SE+\textbf{nu}}, and \textcolor{Goldenrod}{SA-GAN-GCL} respectively. For each curve, we run three times of each algorithm and report the mean and standard-variance, which are plotted in bold and lighter color regions respectively. y-axis values are scores that each is measured as an average of over 100 episodes. x-axis values are episodes.} 
\label{fig:exp_fetch_push}
\vspace{-1em}
\end{figure*}

\begin{figure*}[!t]
\centering
\includegraphics[width=0.88\textwidth]{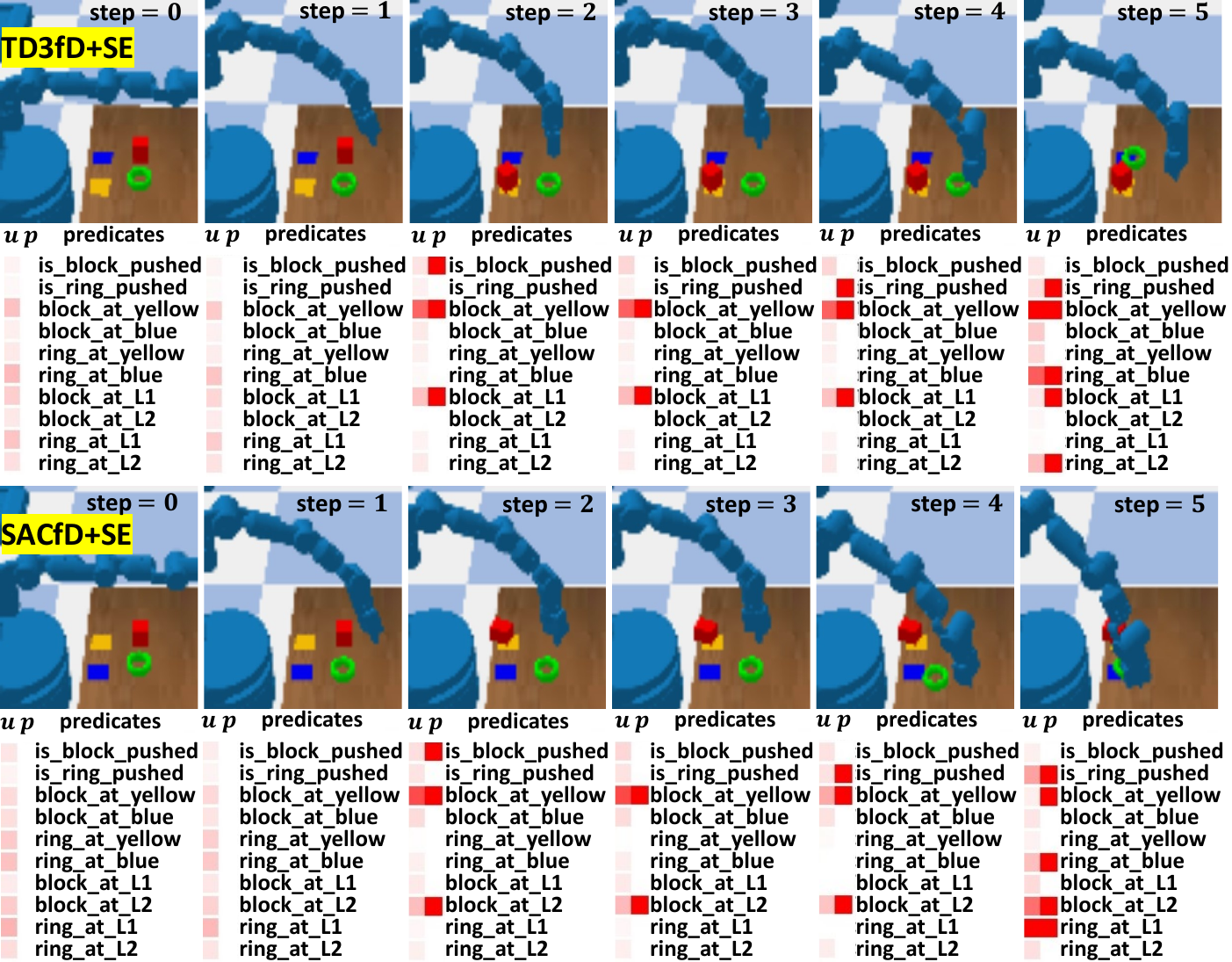}
\caption{\yantian{Examples of predicted self-explanations from agents \textcolor{red}{TD3fD+SE} and \textcolor{red}{SACfD+SE} in the \underline{Robot-Push} domain.}} 
\label{fig:se_vis_robot_push_1_main}
\vspace*{-3mm}
\end{figure*} 

To answer 1, we compare the performance of an RLfD model and the same one supported by our SE-Net. To answer 2, we investigate different ways of using self-explanations -- \textcolor{red}{RLfD+SE}: using both reward and state augmentation, \textcolor{Magenta}{RLfD+SE+\textbf{nu}}: without using the predicate utility weights $u$ to augment RL states $\langle s,p\rangle$, and \textcolor{Aquamarine}{RLfD+SE+\textbf{nrs}}: without adding reward shaping. To answer 3, we investigate a diverse set of RL agents. To answer 4, we compare SERLfD with the SA-GAN-GCL \cite{fu2017learning} (for Imitation Learning). To answer 5, we evaluate our models in both continuous robotic control domains and a discrete Pacman domain. Finally, in our \textbf{supplemental video} and Appendix.D, we show the self-explanations produced by our SE-Net. While the self-explanations are for guiding an RL agent itself, they do reveal some interesting patterns showing robot's understanding of tasks. 

\subsection{Experiments in Continuous Domain} \label{sec:exp_continuous}
\begin{figure*}[!tbh]
\centering
\includegraphics[width=\textwidth]{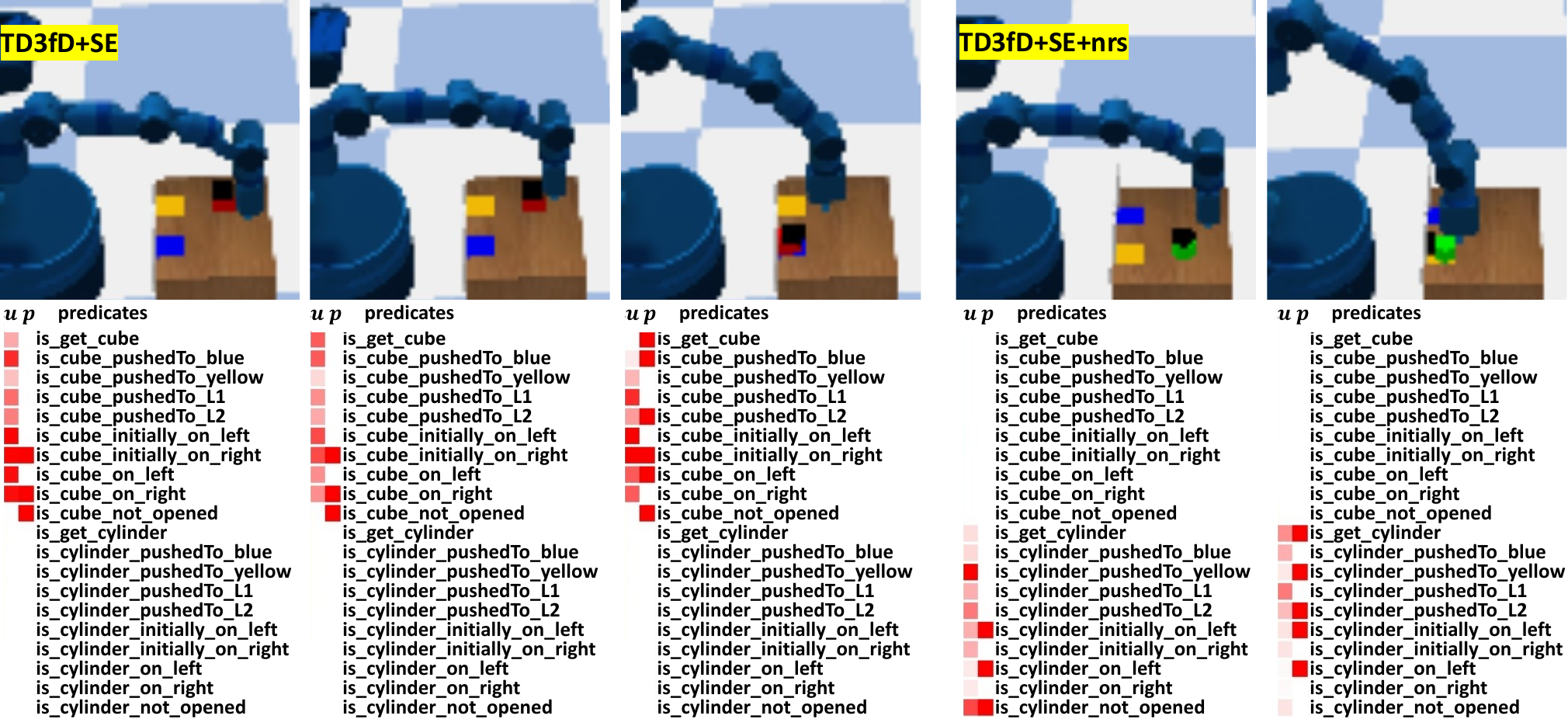}
\caption{\yantian{Examples of predicted self-explanations by agents \textcolor{red}{TD3fD+SE} and \textcolor{red}{SACfD+SE} in the \underline{Robot-Remove-and-Push} domain.}}
\label{fig:se_vis_robot_remove_and_push_main}
\end{figure*}

\mund{Robots and Domains}
We use a Fetch Mobile Manipulator \cite{wise2016fetch} that has a 7-DoF arm in PyBullet simulator \cite{coumans2016pybullet}. We also fixed its mobile base in experiments. We consider continuous robot control domains that are increasingly complex: Robot-Push-Simple, Robot-Push, and Robot-Remove-and-Push. The \underline{Robot-Push} domain (Fig \ref{fig:illustration}.1) has two objects, a block and ring, and two target regions that are colored in yellow and blue. By saying a ``region'', we mean a square that each side is approximately 0.1m. The colors yellow and blue are exchangeable in different episodes. In the \underline{Robot-Push-Simple} domain, however, the colors yellow and blue are always fixed. We assume that there is a roboticist who gives the predicates \{is\_block\_pushed, is\_ring\_pushed, block\_at\_yellow, block\_at\_blue, ring\_at\_yellow, ring\_at\_blue, block\_at\_L1, block\_at\_L2, ring\_at\_L1, ring\_at\_L2\} to the Robot-Push domain (L1 and L2 are the indices of target regions and are always fixed), and gives the predicates \{is\_block\_pushed, is\_ring\_pushed, block\_at\_yellow, block\_at\_blue, ring\_at\_yellow, ring\_at\_blue\} to the Robot-Push-Simple domain. \underline{Robot-Remove-and-Push} (in Fig. \ref{fig:exp_fetch_push}.1) is a robot domain that goes beyond ``Pushing''. A roboticist provides 20 predicates (\yantian{as listed in Fig. \ref{fig:se_vis_robot_remove_and_push_main})}for this domain. Similar to the Robot-Push domain, there are two target regions indexed with L1 and L2 which are fixed and could be assigned with either blue-yellow or yellow-blue colors. In each episode, either a block or a cylinder would show up and both have a removable black cover at the top.

\mund{Tasks}
While each of the above domains and the corresponding background domain knowledge (predicates) support the learning of multiple tasks like pushing an object to a specific region in a specific color or with a specific index, we uniformly assume that human users provide demonstrations for the task of pushing objects (e.g. a block and a ring) into regions in specific colors (yellow or blue), no matter which region is indexed by L1 or L2. In the more complex Robot-Remove-and-Push domain, if the block or cylinder's initial poses are on the left side of the table, the robot needs to push them to their target regions \textit{with the black cover removed}. 

\mund{Settings}
We collected demonstrations by using a keyboard to input control commands. Specifically, we gathered 8 trajectories (average length: 17 steps) for Robot-Push-Simple, 15 trajectories (average length: 19 steps) for Robot-Push, and 16 trajectories (average length: 5 steps) for Robot-Remove-and-Push domains. We use two state-of-the-art RLfD baselines for continuous control tasks: Twin-Delayed DDPG \cite{fujimoto2018addressing} from Demonstrations (TD3fD), and Soft-Actor Critic \cite{haarnoja2018soft} from Demonstrations (SACfD). Each training episode had a maximum of 50 steps. \yantian{For comprehensive details regarding the state space, action space, termination conditions, and reward function, kindly consult Appendix.C in the arXiv version (refer to \cref{arxiv}).}

\mund{Results and Analysis}
The results in Fig. \ref{fig:exp_fetch_push}.1 highlight the effectiveness of RL agents equipped with SE-Nets, particularly \textcolor{red}{TD3fD/SACfD+SE} and \textcolor{Aquamarine}{TD3fD/SACfD+SE+\textbf{nrs}}, which exhibit increased stability and achieve higher scores in a shorter timeframe. Notably, in the Robot-Push-Simple domain with SACfD as the baseline RL agent, the use of self-explanations becomes less crucial due to the lower complexity of the task and the efficacy of entropy-driven exploration. Our investigation focuses on determining the optimal approach for RL agents to leverage self-explanations. The findings emphasize the significance of augmenting states with self-explanations, as this approach provides comprehensive information that enables RL agents to effectively utilize self-explanation details. It is important to note that \textcolor{red}{TD3fD/SACfD+SE}, \textcolor{Magenta}{TD3fD/SACfD+SE+\textbf{nu}}, and \textcolor{Aquamarine}{TD3fD/SACfD+SE+\textbf{nrs}} employ the same SE-Nets, trained in a similar manner, with the key distinction lying in how RL agents utilize self-explanations.

\yantian{
\mund{Visualizing and Analyzing Self-Explanations} We visually inspect the self-explanations generated by our SE-Nets on RL states. Figures \ref{fig:se_vis_robot_push_1_main} and \ref{fig:se_vis_robot_remove_and_push_main} depict the self-explanation predictions in the Robot-Push and Robot-Remove-and-Push domains, respectively. For Robot-Push-Simple and Pacman domains, please refer to Figures 8 and 11 in Appendix.D in the arXiv version (see \cref{arxiv}).

To visualize the self-explanations over time, at each step, we show the original input frame with predicates, their groundings (\textbf{\textit{p}} column), and predicted utility weights (\textbf{\textit{u}} column). Each cell in the \textbf{\textit{p}} columns is either red or white -- meaning whether a predicate is satisfied or not; Each cell in the \textbf{\textit{u}} columns has a color ranging from white to red -- meaning the increasing relevance between the predicate and a successful decision-making that the robot hypothesizes. Each utility weight is normalized by dividing over the sum of all utility weights across a trajectory.

The self-explanations generated by \textcolor{red}{TD3fD+SE} and \textcolor{red}{SACfD+SE}, alongside input frames and predicate groundings, reveal consistent patterns -- They assign higher utility weights to predicates associated with colors rather than predicates related to specific locations. In Fig. \ref{fig:se_vis_robot_push_1_main}, when the block is pushed to the yellow region (note the color exchange of the target regions), the robot demonstrates certainty that ``block\_at\_yellow" is more relevant than ``block\_at\_L1/L2" at step 2. Interestingly, after step 2, since there is only one free target region remaining, the importance of either ``ring\_at\_blue" or ``ring\_at\_L1/L2" becomes inconsequential. Thus, at step 5, \textcolor{red}{SACfD+SE} assumes that ``ring\_at\_L1" is more important than ``ring\_at\_blue", while \textcolor{red}{TD3fD+SE} continues to emphasize the significance of ``ring\_at\_blue".

We now show the produced self-explanations in the domain Robot-Remove-and-Push in Fig. \ref{fig:se_vis_robot_remove_and_push_main}. In the two examples, we can observe that the robot understands which predicates are more task-relevant -- hypothesizing higher utility weights on ``is\_cube\_pushedTo\_blue'' in the \textcolor{red}{TD3fD+SE} case, and ``is\_cylinder\_pushedTo\_yellow'' in the \textcolor{Aquamarine}{TD3fD+SE+\textbf{nrs}} case -- which also guides the following behavior. 

For more visualizations and analysis of self-explanations, including those predicted by trained \textcolor{Magenta}{TD3fD/SACfD+SE+\textbf{nu}} and \textcolor{Aquamarine}{TD3fD/SACfD+SE+\textbf{nrs}} models, please consult our supplemental video\footnote{\color{blue}{https://youtu.be/w5nGYOdVMiA?si=TBvUq3pOpRbca4w4}}.
}

\subsection{Experiments in Discrete Domain}



To evaluate SE-Nets in discrete domains, we conduct experiments in a \underline{Pacman} domain, and the results are depicted in Fig. \ref{fig:exp_fetch_push}.2. In this domain, the Pacman's objective is to eat ghosts within a fixed time frame after consuming a power pellet. The task is considered completed when the Pacman successfully eats all randomly-wandering ghosts, resulting in a reward of 1 and the end of the episode. Otherwise, if the task is not completed, the reward is 0. Two predicate variables, \{ghost\_nearby, eat\_capsule\}, are used to indicate the proximity of a ghost to the Pacman and whether the Pacman has consumed a power pellet. The challenge lies in determining the optimal time to eat the pellet. During the demonstrations, the agent tended to hastily consume the pellet when the ghost happened to be nearby. However, in normal circumstances, it is more advantageous for the agent to wait until the ghost approaches before eating the pellet. For our experiments, we employed Soft Q-Learning \cite{haarnoja2017reinforcement} from Demonstration (SQLfD) as our RL agent. We collected 5 demonstrations, each consisting of an average of 18 steps. In our training, each episode was limited to a maximum of 2000 steps. The results indicate that RL agents equipped with SE-Nets outperform their counterparts in this discrete domain. For further discussions, please refer to Appendix.D and Fig.11 in our arXiv paper (\cref{arxiv}).

\vspace*{-3mm}
\section{Discussion and Future Work} 
\label{sec:conclusion}
Our contribution introduces the SERLfD framework, utilizing self-explanations to elevate the efficacy of RLfD learning by simulating human-like introspection. \yantian{The initial version,  accessible on arXiv in \cref{arxiv}, pioneers the integration of self-explanation into robot learning using deep neural networks.}The SERLfD framework demonstrates substantial advantages through effective handling of ambiguous demonstrations, where only specific underlying factors within each observation pertain to the task. Notably, SERLfD holds potential for practical application given its relevance to real-world scenarios where robots collaborate with non-expert humans. A potential constraint lies in the reliance on optimal object trackers for accurate relation detection, as real-world object tracking and predicate detection can be subject to noise. Future endeavors could investigate the performance of advanced visual models in attaining comparable results and implement our framework on real robotic systems.


\yantian{
Future research can explore a broader range of self-explanation techniques for diverse robot learning challenges. Furthermore, refining self-explanation mechanisms in demanding settings shows promise. Recent strides in using foundation models like LLMs for improved robot learning (see \cite{liu2023reflect}) highlight opportunities for their use in self-explanation-guided robot learning, aiding adaptability to new domains. However, persistent challenges include the inherent difficulty for LLMs to self-explain and discover causally important factors \cite{jin2023can}. Additionally, aligning LLMs' high-dimensional outputs with robot-friendly learning signals remains an ongoing hurdle. Tackling these obstacles is pivotal for the effective incorporation of foundation models into self-explanation-guided robot learning, with potential insights from SERLfD.



\section*{Acknowledgements}
This research was supported by ONR grants N00014-18-1-2442, N00014-18-1-2840, N00014-19-1-2119 and N00014-23-1-2409, and a JP Morgan AI Faculty Research Grant to Kambhampati.


}

\bibliography{sesl}

\appendix
\onecolumn

\section{Common Questions and Our Answers}
In this section, we aim to address some of the common concerns and questions that we have received regarding our work, and hope to address any potential doubts or uncertainties that may arise.

\qund{What are the aim and scope of this work?} 

In this work, our focus is on reinforcement learning from demonstration (RLfD) and our aim is to enhance RLfD through the integration of self-explanations. Therefore, we select the state-of-the-art RLfD methods TD3fD and SACfD as our baseline RL agents for evaluation purposes.

\qund{Are there more common environments that show the benefits of using self-explanations?} 

Consider a scenario where a service robot is assigned the task of delivering two mugs to two human users, referred to as $U_A$ and $U_B$. The mugs are distinguishable by their colors, such as red and blue. The human users are situated on opposite sides of a table, designated as $S_A$ and $S_B$ respectively. In a particular demonstration, a human operator directs the robot to bring the red mug to $S_A$ for $U_A$ and then transport the blue mug to $S_B$ for $U_B$. This demonstration can be interpreted in various ways by the robot. One interpretation is that the red mug is intended for $U_A$ and the blue mug for $U_B$. Alternatively, the robot might understand the demonstration as requiring the red mug to be placed at $S_A$ and the blue mug at $S_B$.

\qund{What if the number of domain predicates is scaled up?} 

We conducted evaluations in three continuous control domains that feature an increasing number of domain predicates. We started with Robot-Push-Simple, which had four objects (ring, cube, yellow region, and blue region) and six predicates. Then, we moved to Robot-Push, which included six objects (ring, cube, yellow region, blue region, region L1, and region L2) and ten predicates. Finally, we evaluated Robot-Remove-and-Push, which involved seven objects (ring, cube, yellow region, blue region, region L1, region L2, and a black cover) and twenty predicates. These evaluations allowed us to explore the effectiveness of our approach in domains with different predicate scales. Our work represents an initial step towards contrastively learning self-explanations from successful and unsuccessful experiences to enhance RLfD. We provide theoretical insights and empirical evidence to support future research in developing practical and adaptable algorithms in this direction.

\qund{Do we need to construct predicates for different tasks?} 

No, we do not assume that we need to construct predicates specifically for each task. Predicates are background domain knowledge that can be applicable to a class of tasks. For instance, in our evaluation domains such as Robot-Push, the predicates can be used to support various tasks, such as pushing a cube and a ring to different locations or pushing them to specific colored regions. While robot experts provide the knowledge of potentially relevant relations, the demonstrations provided by non-expert human users communicate the specific task to the robots. The process for constructing predicates is described in detail in Appendix \ref{sec:det_predicates}.

\qund{How a human expert gives the knowledge of utility functions and reward shaping?} 

In our approach, the knowledge of utility functions and reward shaping is not directly provided by human experts. Instead, we employ a Self-Explanation Network (SE-Net) that learns a utility function by predicting predicate utility weights. The SE-Net is trained using our SERLfD framework, which allows for the self-explanation of demonstrations provided by non-expert human users. Based on the predictions made by the self-explanation, particularly the predicate utility weights, we construct the reward shaping. This approach eliminates the need for human experts to explicitly specify tasks or provide detailed knowledge of utility functions and reward shaping for the robots.

\qund{Why can discriminator training give rise to self-explanations? why the discriminator can be treated as self-explanations?} 

There is a misunderstanding for our technique behind this question. The prediction of utility weights u is accomplished through our Self-Explanation Network (SE-Net). To facilitate GAN-like training, we integrate the SE-Net into a discriminator. The relationship between the SE-Net and discriminator is illustrated in Fig. \ref{fig:SERLfD}. Specifically, the SE-Net predicts utility weights (u) that indicate the importance of predicates in a given state, which can be considered as self-explanations. Although the SE-Net is integrated into the discriminator architecture for training purposes, we exclusively utilize the SE-Net and exclude the discriminator when assisting RL agents in learning tasks based on environment rewards. By leveraging the self-explanations provided by the SE-Net to augment states or rewards, our SERLfD algorithm combines the strengths of RLfD and GAN-IRL, resulting in an effective approach for learning from demonstrations in reinforcement learning settings.

\qund{Is SERLfD simply an extension of SA-GAN-GCL (Fu et al. (2017) \cite{fu2017learning})?} 

No, SERLfD is not simply an extension of SA-GAN-GCL (\cite{fu2017learning}). It can be considered as a combination of RLfD methods and the approach proposed in \cite{fu2017learning}. In our empirical evaluation, we demonstrate that SERLfD harnesses the advantages of both approaches. We compare SERLfD with both state-of-the-art RLfD agents and the work by \cite{fu2017learning} to showcase SERLfD's benefits in learning from demonstrations.

\qund{Why assign -1 to a predicate that is not satisfied, instead of 0?} 

We tested both -1 and 0. We empirically found that \{-1, 1\} works better than \{0, 1\}.

\qund{Why the demonstration shown in Fig. \ref{fig:illustration} is ambiguous?}

The demonstration in Fig. \ref{fig:illustration} illustrates the inherent ambiguity in human demonstrations. When an object is pushed to a target region, both location and color features are detected, leading to the presence of multiple possible explanations. In the example, both ``ring\_at\_blue" and ``ring\_at\_L1" predicates are detected at step 3, but not all of the True predicates might be task-relevant. This ambiguity allows learners to consider either ``ring\_at\_blue", or ``ring\_at\_L1", or both of these predicates as important for the demonstrated task. Similarly, in the Pacman domain, the ambiguity arises in determining the appropriate timing to eat a pellet. The agent in the demonstrations rushed to eat the pellet due to the coincidental proximity of the ghost, whereas the optimal strategy is to wait until the ghost approaches.

\qund{How are demonstrations collected?} 

To collect demonstrations, we use a keyboard to input control commands (Sec. \ref{sec:exp_continuous}).

\qund{How did we determine the success of a trajectory?} 

To determine the success of a trajectory, we employ a reward function comprising a task reward term and a reward shaping term (as described in Equation \ref{eq:serl_rwd}). The reward shaping term relies on utility weights ($u_i$) obtained from a trainable Self-Explanation Network, which essentially provides self-explanation. The task reward function within the RL environment serves as an indicator of whether a trajectory accomplishes the task or not. By leveraging a set of symbolic predicates as human-related background knowledge, our SERLfD framework is designed to excel in more complex sequential decision settings that involve both tacit and explicit task knowledge \cite{kambhampati2022symbols}. The development of success indicators for explicit tasks is relatively straightforward in this context.

\qund{Where do we explain our task settings in this paper?} 

We elaborate our task settings in multiple sections of the paper. In the introduction, we introduce the tasks and their settings using an example to illustrate the scenario. In the evaluation section, we provide detailed explanations of the tasks, including a paragraph specifically dedicated to describing the tasks, and another paragraph that outlines the settings. Additionally, for more comprehensive information about our task settings, we provide additional details in Appendix \ref{sec:settings}.

\qund{For TD3FD+SE+nrs/TD3FD+SE+nu, what does ``nrs'' and ``nu'' stand for?} 

As mentioned in Section \ref{sec:eval}, ``nrs" stands for ''no reward shaping", and ``nu" stands for ``no utility weights". These refer to the two different ways of using self-explanations (predicate utility weights) to support RL agents. In the case of ``RLfD+SE+nrs", we append the predicate utility weights to the state feature (raw state and predicate grounding values) without using reward shaping. This approach involves state augmentation, where the self-explanations guide RLfD agents. On the other hand, ``RLfD+SE+nu" uses the predicate utility weights to compute reward shaping based on Equation \ref{eq:serl_rwd}. This approach involves reward augmentation, where the self-explanations guide RLfD with the aid of reward shaping.

\qund{How do we evaluate other than showing score curves?} 

In addition to presenting score curves, we provide visualizations and analyses of generalized explanations in our supplemental video and Appendix. \ref{sec:appen_vis}. In the video, we comment on the generated self-explanations, and we compare self-explanations from both successful and unsuccessful trajectories that are generated by different models. These visualizations and comparisons allow for a more in-depth understanding of the self-explanation generation process and its impact on the performance of the models. 

\qund{Why we do not compare with GAIL (in short of Generative Adversarial Imitation Learning)?} 

We have already compared our method with SA-GAN-GCL (\cite{fu2017learning}), which is a baseline approach that uses intrinsic rewards from a discriminator to train RL agents. SA-GAN-GCL has been demonstrated to perform comparably to GAIL (as mentioned in the second paragraph of Section 7 in \cite{fu2017learning}), another method that utilizes a discriminator for providing intrinsic rewards. Therefore, by evaluating our method against SA-GAN-GCL, which serves as a representative of this class of methods, we indirectly establish a comparison with GAIL as well. This allows us to assess the effectiveness of our approach in a similar context to GAIL.

\qund{Are the comparisons fair between models that use and do not use self-explanations?} 

Yes, the comparisons between models that use self-explanations and those that do not are fair. In all of our RL agents, including both models that use self-explanations and models that do not, we append the predicate grounding values (an array of values 1 or -1) to the raw state feature. This ensures that all models have access to the same information (especially regarding the predicate grounding), allowing for a fair comparison of their performance. This clarification is provided in Section \ref{sec:ap} of our work.

\qund{Can an experiment be run where the ambiguity in the demonstrations is drastically reduced to see whether SE still helps then?} 

Yes, an experiment can be conducted to drastically reduce the ambiguity in the demonstrations and observe the impact of self-explanation in such a scenario. In our work, we have actually designed the Robot-Push-Simple domain specifically to explore the performance of SERLfD when there is minimal ambiguity. In this domain, only color-relevant predicates are present, and there are no location-relevant predicates. As a result, robots only need to self-explain with respect to color-related relations. Even in this setting, our results clearly demonstrate that self-explanation still benefits the TD3fD agent, indicating the effectiveness of our approach.

\qund{The standard deviation of the compared methods (Fig. \ref{fig:exp_fetch_push}.1) is very high which makes it difficult to correctly assess the superiority of the proposed methods} 

\begin{figure*}[!h]
\centering
\includegraphics[width=1.\columnwidth]{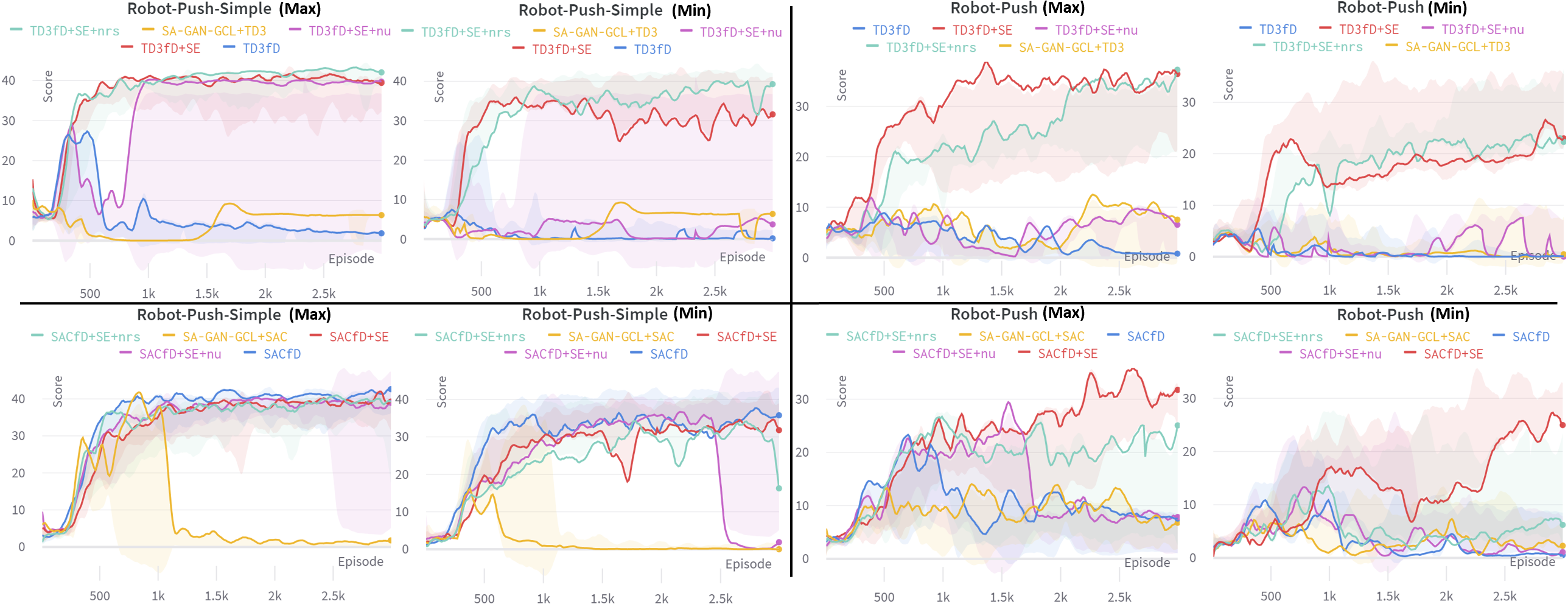}
\caption{Learning curves of training the baseline RL agents (TD3fD/SACfD), RL agents with Self-Explainer (TD3fD+SE/SACfD+SE), the ablation studies of removing utility weights or reward shaping from RL agent's inputs or rewards respectively: TD3fD+SE+\textbf{nu}/SACfD+SE+\textbf{nu} or TD3fD+SE+\textbf{nrs}/SACfD+SE+\textbf{nrs}, and an Imitation Learning agent built by using RL in the original SA-GAN-GCL framework \cite{fu2017learning}. For each curve we run three times of each algorithm and report the \textbf{Max} and \textbf{Min} values. The lighter color region shows standard-deviation. y-axis values are scores that each is measured as an average over 100 episodes. x-axis values are episodes and each has at most 50 steps.} 
\label{fig:exp_fetch_push_maxmin_1}
\end{figure*}

We acknowledge that the standard deviation of the compared methods in Fig. \ref{fig:exp_fetch_push}.1 is relatively high. However, this high variability mainly arises from training instabilities observed in some of the ablation models (compared methods). To clearly show the effectiveness of our winner models with using self-explanations, RLfD+SE and RLfD+SE+nrs, we have included additional results in  Fig. \ref{fig:exp_fetch_push_maxmin_1}. These figures display the minimum and maximum values of the score curves, allowing for a better understanding of the performance range of each method. By examining these curves, it becomes evident that our winning models, RLfD+SE and RLfD+SE+nrs, consistently outperform the other models in terms of both stability and overall performance.

\section{Detecting Predicates Values} \label{sec:det_predicates}

In the field of Robot Learning from Demonstration (LfD), some works have utilized marker tracking to simplify the vision component of their systems, as shown in studies such as \cite{konidaris2012robot} and \cite{niekum2012learning}. Inspired by these approaches, we propose a procedural method for determining the truth values of predicates. We assume that our domain consists of fully observable objects that can be tracked using an object-tracking module. This module provides the bounding box and current pose of each tracked object. In our work, we extract the current poses of all domain objects directly from the PyBullet simulator \cite{coumans2016pybullet}. To evaluate whether a state, represented by the tracking results, satisfies a grounded predicate, we employ pre-defined rules. It is important to clarify that our primary focus is not on recognizing object relations from complex sensory data. Instead, we leverage the object-tracking module and predefined rules to compute the truth values of predicates in our domain.

To determine the truth values of predicates, we have predefined rules for different types of predicates. These rules help classify the poses of tracked objects and determine whether they satisfy the predicate variables.

For the predicate variable ``is\_X\_pushed'', we compute its predicate value by checking if there is a change in the relative pose of object X to the world by more than 0.03 meters. In our domain, the object X is placed on a table and would only move due to external forces.

For the predicate variables ``X\_at\_yellow'' or ``X\_at\_blue'', we detect their predicate values by checking if the $(x, y)$ values of the object X's pose fall within the yellow or blue region, which is defined as a square with dimensions of $0.1m \times 0.1m$. Additionally, the $z$ value of X's pose should not change significantly, with a tolerance of 0.075 meters.

For the predicate variables ``X\_at\_L1" or ``X\_at\_L2," we determine their predicate values by checking if the center of the $(x, y)$ values of the object X's pose falls within a $0.1m \times 0.1m$ square with either L1 or L2 as the center. Similarly, the $z$ value should not change significantly, with a tolerance of 0.075 meters.

By applying these predefined rules, we can classify the poses of tracked objects and determine the truth values of the corresponding predicate variables.


\section{Additional Evaluation Setting Details} \label{sec:settings}
In this section, we provide more information about our evaluation settings. 

\subsection{Robotics Experiments}

\begin{figure}[!h]
\centering
\includegraphics[width=11cm]{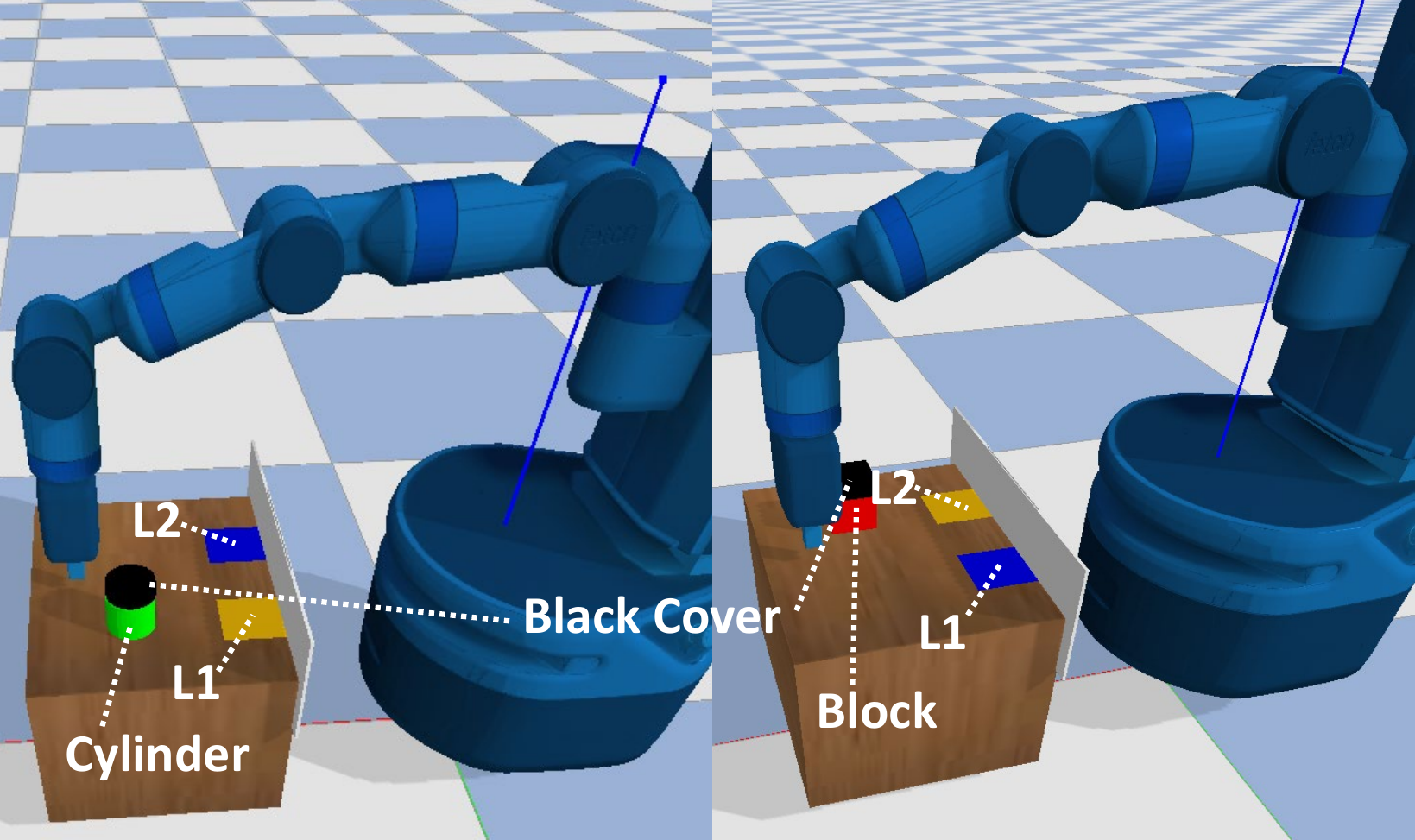} 
\caption{The Robot-Remove-and-Push Domain}
\label{fig:domain-1}
\end{figure}

\subsubsection{State Space}
Our state-space design follows the fetch-push environment implemented in OpenAI Gym \footnote{\url{https://gym.openai.com/envs/FetchPush-v0/}}. It includes the positions and orientations of various objects within the domain with respect to the world frame. These objects, such as the end-effector, ring, block, regions in blue and yellow colors, and regions L1 and L2, can be tracked using a state-of-the-art visual tracker. To enable a fair comparison between RLfD with our Self-Explainers and other baseline models, we append the predicate grounding values to this state feature.

\subsubsection{Action Space}
The action space in our framework is defined as a 4-tuple: [translation\_x, translation\_y, translation\_z, yaw\_angle] of the end-effector with respect to the world frame. This definition aligns with the work of Quillen et al. \cite{quillen2018deep}.

\subsubsection{Termination Condition} \label{sec:settings:termination}
In our RL environments, the termination conditions are designed to determine whether an episode is a Success or a Failure. In the Robot-Remove-and-Push domain, each episode involves either a block or a cylinder with a black cover. The robot's objective is to push the block to the blue region or the cylinder to the yellow region. If the block/cylinder is initially located on the left part of the table, the robot must also remove the black cover before completing the task. In the Robot-Push and Robot-Push-Simple domains, success is achieved when the block is successfully pushed to the yellow region and the ring is pushed to the blue region. However, the RL environment can terminate with a Failure under two conditions: 

(a) If any object to be pushed, whether it's the ring/cylinder or the block, falls off the table during the episode.

(b) If the maximum time step is reached before the robot successfully completes the task.

These termination conditions allow the RL environment to determine whether an episode is a Success or a Failure based on the accomplishment of the objectives and other specified criteria.

\subsubsection{Reward Function}

In our evaluation, we consider both sparse and non-sparse reward settings to thoroughly assess the benefits of using self-explanations. The Robot-Remove-and-Push and Pacman domains (explained in Section \ref{sec:pacman:rwds}) are designed with sparse rewards, while the Robot-Push-Simple and Robot-Push domains have less sparse rewards. Our experiments demonstrate that incorporating self-explanations improves performance in both sparse and less sparse reward settings, indicating the uniform benefits of using self-explanations.

In the Robot-Remove-and-Push domain, the robot is rewarded with +50 only when it successfully accomplishes the task. For the Robot-Push and Robot-Push-Simple domains, the reward function is designed to be less sparse. In most situations, the reward $r(s,a)$ is zero. However, if the robot pushes an object and brings it $\delta d$ closer to its target location, it receives a reward of $100*\delta d$. Additionally, if either the ring or the block is successfully pushed into its target region, the robot receives a reward of +25. If both the ring and the block are pushed into their respective target regions, the robot receives a reward of +50.

\begin{figure*}[!t]
\vspace*{-2mm}
\centering
\includegraphics[width=\textwidth]{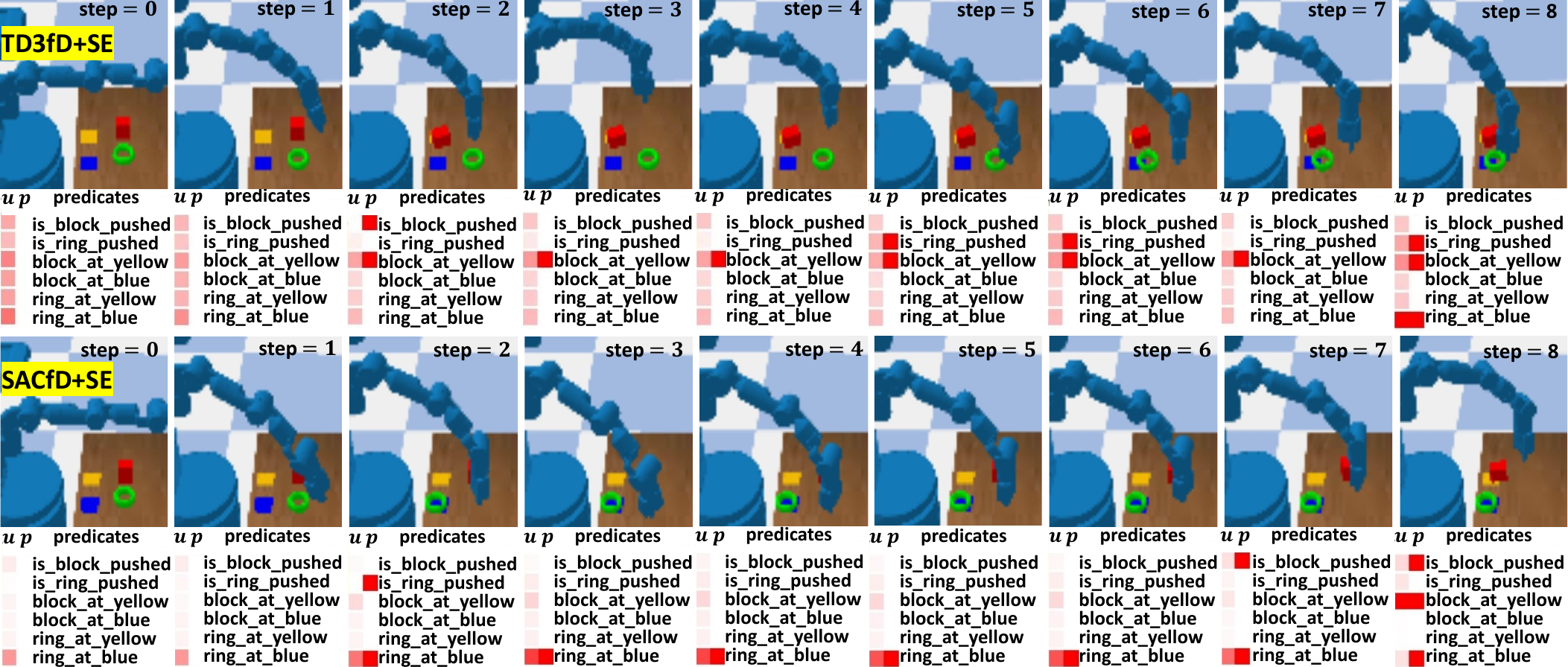}
\caption{Visualizing the predicted self-explanations from agents TD3fD+SE and SACfD+SE in Robot-Push-Simple domain.} 
\label{fig:se_vis_robot_push_simple_1}
\end{figure*} 

\begin{figure}[!tbh]
\centering
\includegraphics[width=1.0\columnwidth]{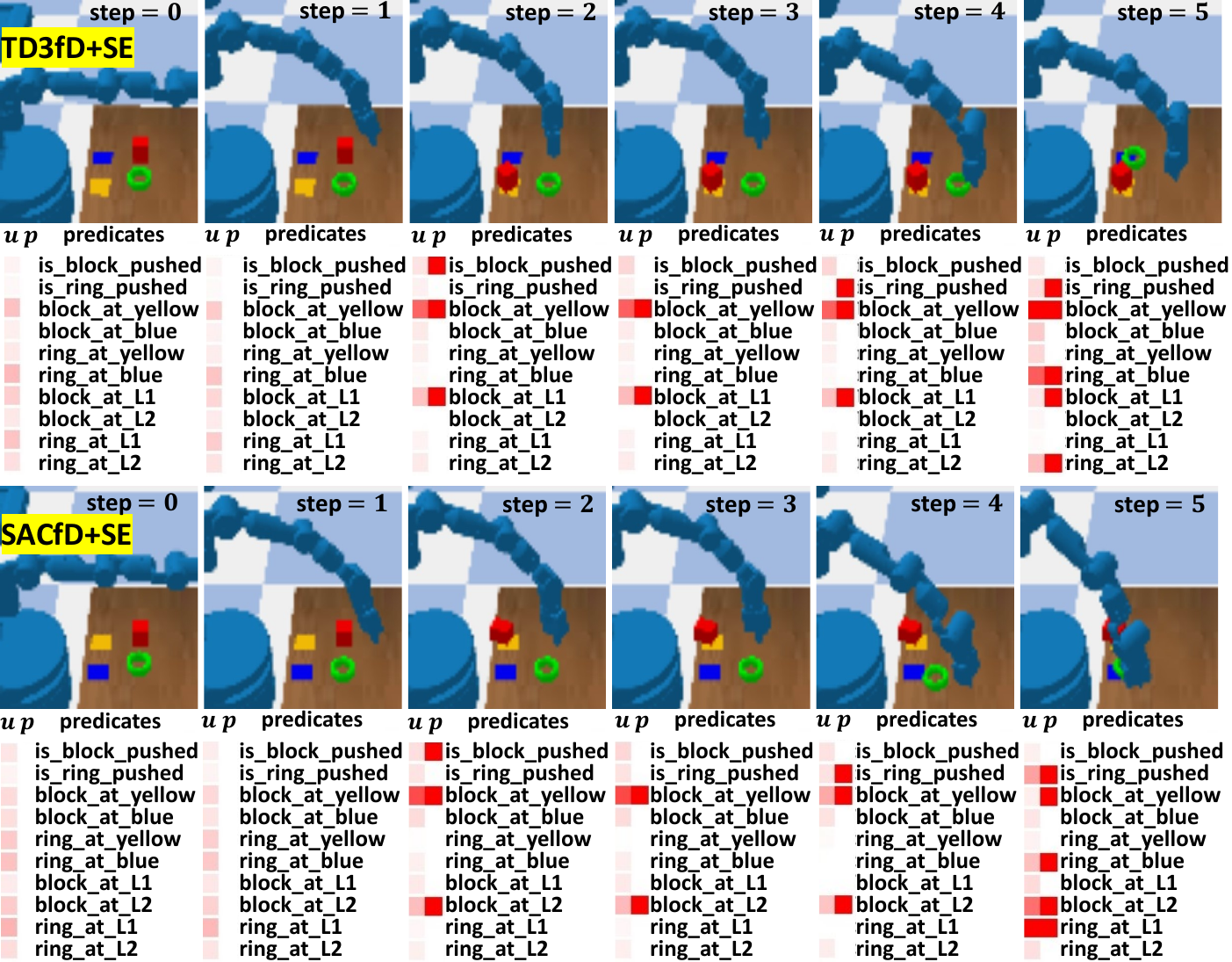}
\caption{Visualizing the predicted self-explanations from agents TD3fD+SE and SACfD+SE in Robot-Push domain.} 
\label{fig:se_vis_robot_push_1}
\vspace*{-6mm}
\end{figure} 

\begin{figure*}[!tbh]
\centering
\includegraphics[width=\textwidth]{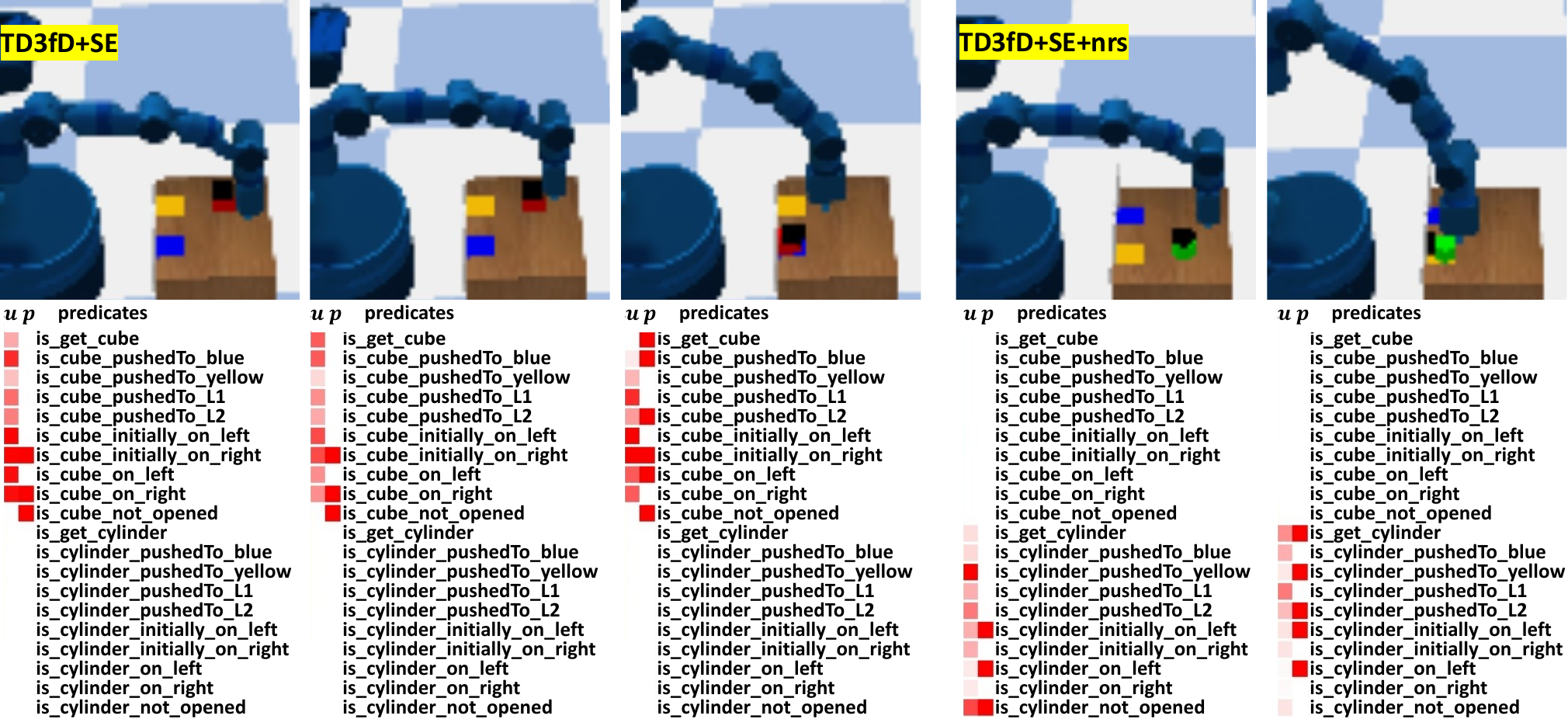}
\caption{We visualize the self-explanations predicted by the agents TD3fD+SE and TD3fD+SE+nrs in Robot-Remove-and-Push domain. We analyze the visualizations in Sec. \ref{sec:appen_vis}.} 
\label{fig:se_vis_robot_remove_and_push}
\vspace*{-1mm}
\end{figure*} 

\begin{figure*}[!tbh]
\centering
\includegraphics[width=\textwidth]{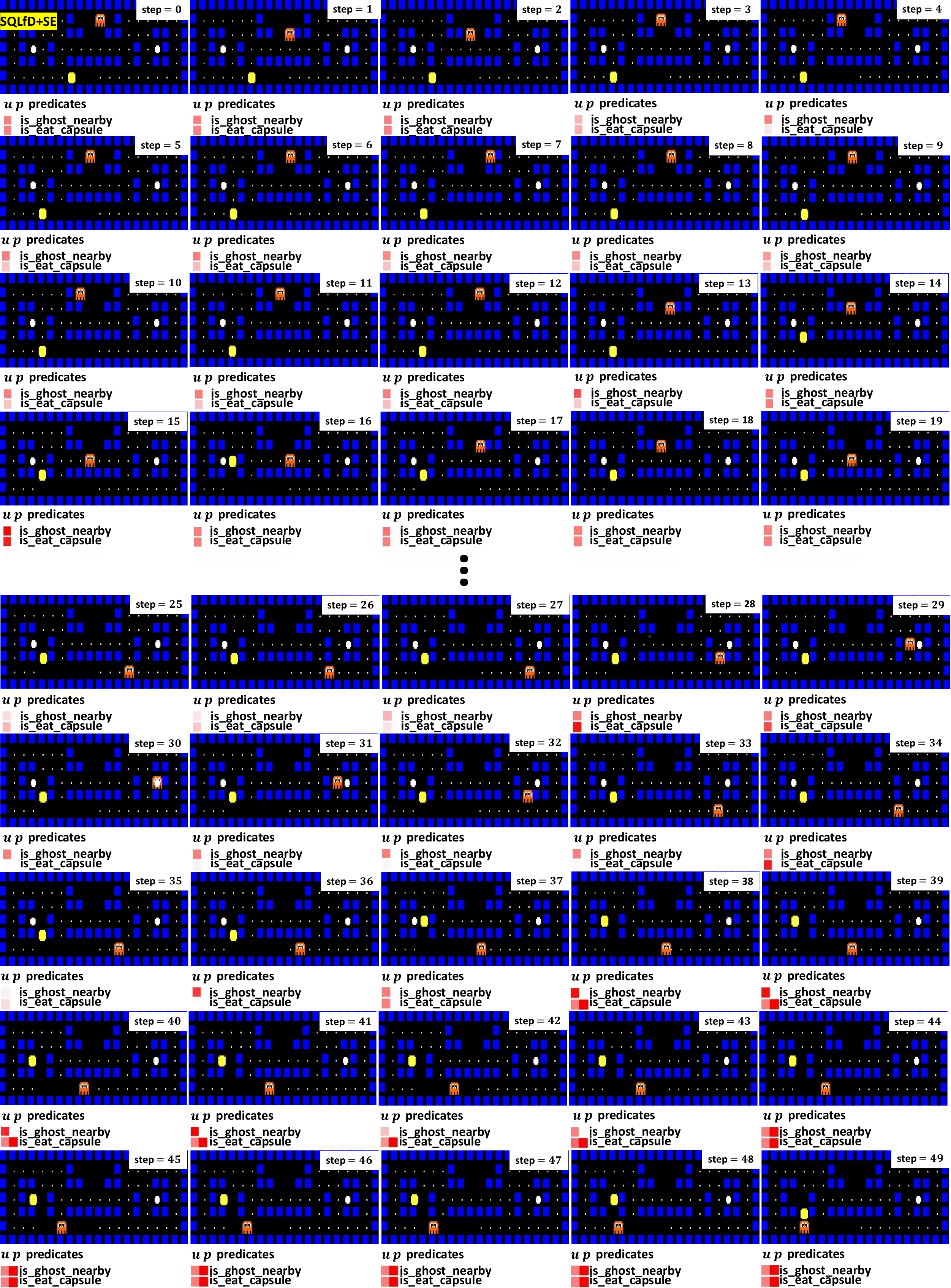}
\caption{We visualize the self-explanations predicted by the agent SQLfD+SE, together with the original input frames in Pacman domain. We analyze the visualizations in Sec. \ref{sec:appen_vis}.} 
\label{fig:se_vis_pacman}
\vspace*{-1mm}
\end{figure*} 
\subsubsection{Robot-Remove-and-Push Predicates} \label{sec:settings:rrap_predicates}
This domain supports a more complex task of 20 predicates.
The 20 predicates are: \{is\_get\_cube, is\_cube\_pushedTo\_blue, is\_cube\_pushedTo\_yellow, is\_cube\_pushedTo\_L1,
                      is\_cube\_pushedTo\_L2, is\_cube\_initially\_on\_left, is\_cube\_initially\_on\_right, is\_cube\_on\_left,
                      is\_cube\_on\_right, is\_cube\_not\_opened, is\_get\_cylinder, is\_cylinder\_pushedTo\_blue, is\_cylinder\_pushedTo\_yellow, is\_cylinder\_pushedTo\_L1, is\_cylinder\_pushedTo\_L2, is\_cylinder\_initially\_on\_left,
                      is\_cylinder\_initially\_on\_right, is\_cylinder\_on\_left, is\_cylinder\_on\_right, is\_cylinder\_not\_opened\}

\subsection{Pacman Experiment}
\subsubsection{State Space}
In the Pacman experiments, the state space comprises RGB-image observations. To capture temporal information, a common practice in Deep Reinforcement Learning is to stack $k$ consecutive image frames together, forming a Markov Decision Process (MDP) state. In our Pacman experiments, we set $k$ to 3, meaning that three consecutive frames are stacked to represent the current state. 

To facilitate a fair comparison between RLfD with our Self-Explainers and other baseline models, we append the predicate grounding values to this state feature. This allows for a consistent representation across different approaches and ensures that the Self-Explainers' performance can be evaluated in the same context as the baseline models.

\subsubsection{Action Space}
The action space in the Pacman environment consists of five allowed actions:

\begin{enumerate}
    \item \texttt{move-up}: Moves the Pacman agent one step upwards.
    \item \texttt{move-down}: Moves the Pacman agent one step downwards.
    \item \texttt{move-left}: Moves the Pacman agent one step to the left.
    \item \texttt{move-right}: Moves the Pacman agent one step to the right.
    \item \texttt{no-op}: No operation, which means the Pacman agent remains in the same place.
\end{enumerate}

These actions provide the agent with the ability to navigate and interact with the environment to achieve its objectives.

\subsubsection{Termination Condition} \label{sec:settings:termination_pacman}
In the Pacman environment, an episode will terminate with success when the Pacman eats power pellets and successfully consumes all the ghosts while the power pellets are in effect. The power pellets grant the Pacman the ability to devour the ghosts temporarily, and if the Pacman can eliminate all the ghosts within this time frame, the episode ends successfully.

On the other hand, the episode will terminate with failure if either of the following conditions is met:

(a) The Pacman is eaten by a ghost. This occurs when the Pacman collides with a ghost without having consumed an effective power pellet. If the Pacman fails to defend against the ghost and gets eaten, the episode terminates as a failure.

(b) The maximum time step is reached. Each episode has a predefined maximum number of steps it can take. If the Pacman has not eliminated all the ghosts within this limit, the episode will terminate as a failure due to reaching the maximum allowed time.

These termination conditions dictate the outcome of each episode in the Pacman environment, determining whether it concludes with success or failure based on the Pacman's actions and interactions with the ghosts, taking into consideration the availability and duration of power pellets.

\subsubsection{Reward Function} \label{sec:pacman:rwds}
In the Pacman environment, we utilize a sparse binary reward function that assigns a reward of +1 only when the Pacman agent successfully consumes all the ghosts. This means that the reward is binary in nature, with a positive reward given for the successful completion of the task and a zero reward for all other actions or states.

\section{Visualizing and Analyzing Self-Explanations} \label{sec:appen_vis}

In this section, we visualize and analyze the self-explanations produced by running our SE-Nets on RL states. To visualize the self-explanations over time, at each step, we show the original input frame with predicates, their groundings (\textbf{\textit{p}} column), and predicted utility weights (\textbf{\textit{u}} column). Each cell in the \textbf{\textit{p}} columns is either red or white -- meaning whether a predicate is satisfied or not; Each cell in the \textbf{\textit{u}} columns has a color ranging from white to red -- meaning the increasing relevance between the predicate and a successful decision-making that the robot hypothesizes. Each utility weight is normalized by dividing over the sum of all utility weights across a trajectory.

The self-explanations generated by TD3fD+SE and SACfD+SE are visualized alongside input frames and predicate groundings in Fig. \ref{fig:se_vis_robot_push_simple_1} (Robot-Push-Simple) and Fig. \ref{fig:se_vis_robot_push_1} (Robot-Push). Upon examination, we observe that both TD3fD+SE and SACfD+SE exhibit similar and logical patterns in their self-explanations. They assign higher utility weights to predicates associated with colors rather than predicates related to specific locations. In Fig. \ref{fig:se_vis_robot_push_simple_1}, TD3fD+SE assigns higher weights to ``block\_at\_yellow" while SACfD+SE assigns higher weights to ``ring\_at\_blue". These self-explanations guide the RL agents to learn the optimal behavior, where TD3fD+SE first pushes the block to the yellow region, and SACfD+SE first pushes the ring to the blue region. Importantly, both approaches achieve equally optimal outcomes, as the order of object pushing does not affect task completion. At step 2, when the block and ring are successfully pushed to the yellow and blue regions, respectively, TD3fD+SE and SACfD+SE become more certain about the importance of ``block\_at\_yellow" and ``ring\_at\_blue". Similarly, in Fig. \ref{fig:se_vis_robot_push_1}, when the block is pushed to the yellow region (note the color exchange of the target regions), the robot demonstrates certainty that ``block\_at\_yellow" is more relevant than ``block\_at\_L1/L2" at step 2. Interestingly, after step 2, since there is only one free target region remaining, the importance of either ``ring\_at\_blue" or ``ring\_at\_L1/L2" becomes inconsequential. Thus, at step 5, SACfD+SE assumes that ``ring\_at\_L1" is more important than ``ring\_at\_blue", while TD3fD+SE continues to emphasize the significance of ``ring\_at\_blue".

We now show the produced self-explanations in the domain Robot-Remove-and-Push in Fig. \ref{fig:se_vis_robot_remove_and_push}. In the two examples, we can observe that the robot understands which predicates are more task-relevant -- hypothesizing higher utility weights on ``is\_cube\_pushedTo\_blue'' in the TD3fD+SE case, and ``is\_cylinder\_pushedTo\_yellow'' in the TD3fD+SE+nrs case -- which also guides the following behavior. 

Finally in the Pacman domain, Fig. \ref{fig:se_vis_pacman} shows the self-explanations predicted from a Soft Q-Learning with SE-Net (SQLfD+SE) agent. In the beginning, e.g. from the step 0 to step 13, the Pacman pays more attention to how close the ghost is, since the Pacman frequently hypothesizes a higher utility weight to the predicate ``is\_ghost\_nearby''. This is reasonable because there is a duration for the power capsule to be effective (which allows the Pacman to eat a ghost). After eating a capsule and before the duration ends, the Pacman has to eat the ghost. When the ghost approaches to the Pacman, or the ghost is not too far away from the Pacman, the SE-Net tends to give the predicate ``is\_eat\_capsule'' a higher utility weight. For example, from step 0 to 3, the ghost is not too far away from Pacman. From step 4 to 13, the ghost seems to move further and Pacman does not care about the predicate ``is\_eat\_capsule'' too much. From step 14 to step 19, the ghost shows a tendency of approaching the Pacman and thus we observe that ``is\_eat\_capsule'' has high utility weights across those steps. Likewise, at step 34, the ghost approaches again to the Pacman and the Pacman eventually decides to eat a capsule at step 38. After the step 38, the Pacman wants to eat the ghost and therefore we can see that the Pacman believes both ``is\_eat\_capsule'' and ``is\_ghost\_nearby'' are important.

We now show the produced self-explanations in the domain Robot-Remove-and-Push in Fig. \ref{fig:se_vis_robot_remove_and_push}. In the two examples, we can observe that the robot understands which predicates are more task-relevant -- hypothesizing higher utility weights on ``is\_cube\_pushedTo\_blue'' in the TD3fD+SE case, and ``is\_cylinder\_pushedTo\_yellow'' in the TD3fD+SE+nrs case -- which also guides the following behavior.

For the self-explanation visualization of other ways of using self-explanations (RLfD+SE+nu and RLfD+SE+nrs), we encourage readers to refer to our supplemental video. 

\section{Hyperparameters}
In this section, we provide the values of crucial hyperparameters as listed in Table. \ref{tbl_result}.

\hfill 
\hfill 
\hfill 
\hfill 

\hfill 
\hfill 
\hfill 
\hfill 

\hfill 
\hfill 
\hfill 
\hfill

\hfill 
\hfill 
\hfill 
\hfill 

\begin{table}[h]
\begin{center} {\footnotesize
\begin{tabular}{cc}
\hline
 Parameter 
 & Value\\
\hline
batch size & 64 \\[0ex]
replay buffer size & 200,000 \\[0ex]
learning rates for actor and critic & 0.0003 \\[0ex]
learning rates for self-explainer & 0.001 \\[0ex]
discount ($\gamma$) & 0.99 \\[0ex]
exploration noise & 0.1 \\[0ex]
minimal exploration noise & 0.005 \\[0ex]
noise decay period & 5,000 \\[0ex]
initial random actions & 5,000 \\[0ex] 
RLfD pretrain steps & 200 \\[0ex] 
initial random actions & 5,000 \\[0ex] 
number of hidden layers (all networks) & 2 \\[0ex] 
number of hidden units for layer 1 and 2 & [400, 300] \\[0ex] 
nonlinearity & ReLU \\[0ex] 
prioritized experience replay $\alpha$ value & 0.3 \\[0ex]
prioritized experience replay $\beta$ value & 1 \\[0ex]
prioritized experience replay $\epsilon$ value & 0.000001 \\[0ex]
seeds & We used different random seeds, e.g. 777, 92,450,164, 41,786 \\[0ex]

\hline
\end{tabular} }
\end{center}
\caption{}
\label{tbl_result}
\end{table}


\end{document}